\documentclass{llncs}
\usepackage{graphicx}
\usepackage[utf8]{inputenc}
\usepackage{amsmath}%
\usepackage{subcaption}
\usepackage{fontawesome}
\usepackage{booktabs}
\usepackage{soul}  
\usepackage{graphicx}
\usepackage[utf8]{inputenc}
\usepackage{color}
\usepackage{latexsym}
\usepackage{comment}
\usepackage{enumitem}
\usepackage{hyperref}
\usepackage[colorinlistoftodos]{todonotes}
\usepackage{multirow}
\usepackage{array}
\usepackage{color}
\usepackage{xcolor}
\usepackage{soul}

\usepackage{xspace, colortbl}
\definecolor{Gray}{gray}{0.9}

\newcolumntype{+}{>{\global\let\currentrowstyle\relax}}
\newcolumntype{^}{>{\currentrowstyle}}
\newcommand{\rowstyle}[1]{\gdef\currentrowstyle{#1}%
#1\ignorespaces
}

\newcommand{\ct}{\texttt{CheckThat!}\,}
\newcommand{\politifact}{\textit{PolitiFact}\,}
\newcommand{\ctend}{\texttt{CheckThat!}}

\begin{document}
\title{Overview of CheckThat! 2020: \\Automatic Identification and 
Verification \\ of Claims in Social Media}

%
\titlerunning{Overview of the CLEF-2020 CheckThat! Lab}
%
\author{%
Alberto Barr\'{o}n-Cede\~no\inst{1}
\and 
Tamer Elsayed\inst{2}
\and 
Preslav Nakov\inst{3}	
\and  \\
Giovanni Da San Martino\inst{3}
\and 
Maram Hasanain\inst{2}
\and 
Reem Suwaileh\inst{2}
\and \\
Fatima Haouari\inst{2}
\and
Nikolay Babulkov\inst{4}
\and 
Bayan Hamdan\inst{5}
\and 
Alex Nikolov\inst{4}
\and\\
Shaden Shaar\inst{3}
\and
Zien Sheikh Ali\inst{2}
}

\authorrunning{A. Barrón-Cedeño et al.}

\institute{%
DIT, Università di Bologna, Forlì, Italy \\
\email{a.barron@unibo.it}	\and
Computer Science and Engineering Department, Qatar University, Doha, Qatar \\
\email{\{telsayed, maram.hasanain, rs081123, 200159617, zs1407404\}@qu.edu.qa}\and
Qatar Computing Research Institute, HBKU, Doha, Qatar \\
\email{\{pnakov, gmartino, sshaar\}@hbku.edu.qa}\and
FMI, Sofia University ``St Kliment Ohridski'', Bulgaria\\
\email{\{nbabulkov, alexnickolow\}@gmail.com}\and
Independent Researcher\\
\email{bayan.hamdan995@gmail.com}
}

\maketitle              
\setcounter{footnote}{0}
\begin{abstract}
We present an overview of the third edition of the \ct\ Lab at CLEF 2020. The 
lab featured five tasks in two different languages: English and Arabic. The 
first four tasks compose the full pipeline of claim verification in social 
media:
Task~1 on check-worthiness estimation,
Task~2 on retrieving previously fact-checked claims,
Task~3 on evidence retrieval, and 
Task~4 on claim verification. 
The lab is completed with Task~5 on check-worthiness estimation in political debates and speeches.
A total of 67 teams registered to participate in the lab (up from 47 at CLEF 
2019), and 23 of them actually submitted runs (compared to 14 at CLEF 2019). 
Most teams used deep neural networks based on BERT, LSTMs, or CNNs, and achieved sizable improvements over the baselines on all tasks.
Here we describe the tasks setup, the evaluation results, and a summary of the approaches used by the participants, and we discuss some lessons learned.
Last but not least, we release to the research community all datasets from the lab as well as the evaluation scripts, which should enable further research in the important tasks of check-worthiness estimation and automatic claim verification.
\end{abstract}


\section{Introduction}
\label{sec:intro}

The \ct lab%
\footnote{\url{https://sites.google.com/view/clef2020-checkthat/}}
was run for the third time in the framework of CLEF 2020. The purpose of the 2020 edition was to foster the development of technology that would enable the 
(semi-)automatic verification of claims posted in social media, in 
particular \emph{Twitter}.\footnote{The 2018 edition~\cite{clef2018checkthat} focused on the 
identification and verification of claims in political debates. Beside 
political debates, the 2019 
edition~\cite{ecir-checkthat:2019,clef-checkthat:2019}
also focused on isolated claims in conjunction with a closed set of Web 
documents to retrieve evidence from.}
We turn our attention to Twitter because information posted on that platform is 
not checked by an authoritative entity before publication and such information 
tends to disseminate very quickly.\footnote{Recently, Twitter started flagging some tweets that violate its policy.} Moreover, social media posts lack context due 
to their short length and conversational nature; thus, identifying a claim's 
context is sometimes key for enabling effective 
fact-checking~\cite{cazalens2018content}.

\begin{figure}[t]
\centering
\includegraphics[width=\columnwidth]{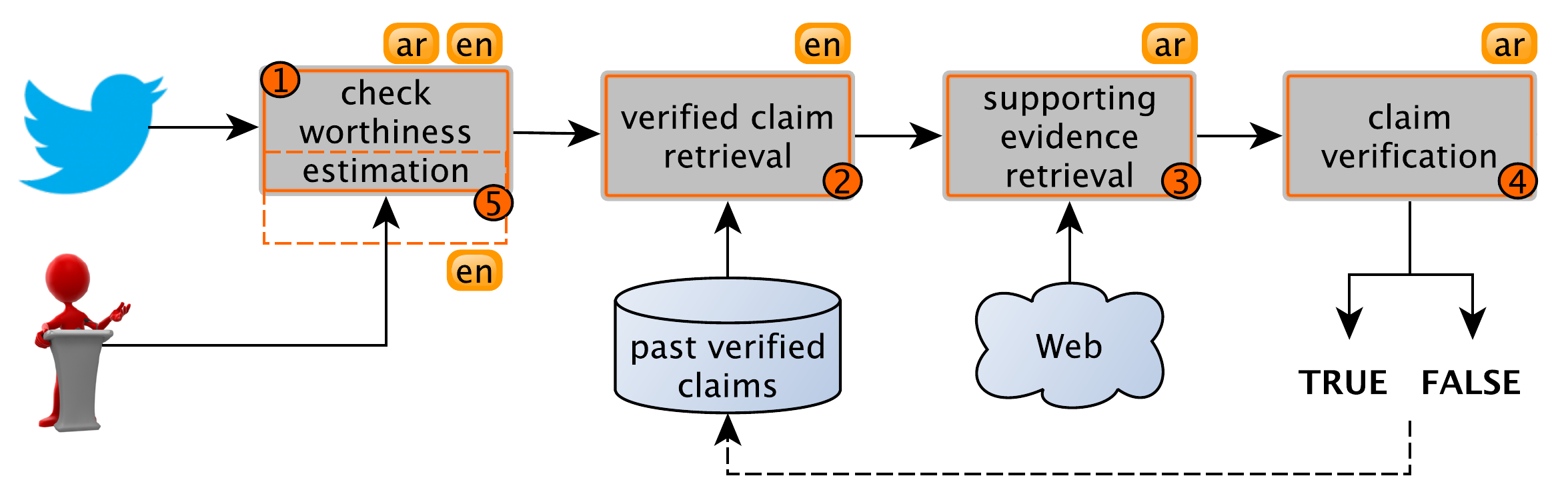}
\caption{The \ct claim verification pipeline. Our tasks cover all four steps of the pipeline in Arabic or English. Tasks 1--4 focus on Twitter, while task 5 is run on political debates and speeches.}
\label{fig:pipeline}
\end{figure}

The full identification and verification pipeline is displayed in 
Figure~\ref{fig:pipeline}. The four tasks are defined as follows:

\begin{description}
 \item[Task 1] Check-worthiness estimation for tweets. Predict which tweet from a stream of tweets on a topic should be prioritized for fact-checking.
 \item[Task 2] Verified claim retrieval: Given a check-worthy tweet, and a set of previously-checked claims, determine whether the claim in the tweet has been fact-checked already.
 \item[Task 3] Evidence retrieval. Given a check-worthy claim in a tweet on a specific topic and a set of text snippets extracted from potentially-relevant webpages, return a ranked list of evidence snippets for the claim.
 \item[Task 4] Claim verification. Given a check-worthy claim in a tweet and a set of potentially-relevant Web pages, estimate the veracity of the claim.
\end{description}

\textbf{Task~5} complements the lab. It is as Task~1, but on political debates ad speeches rather than on tweets: given a debate segmented into sentences, together with speaker information, prioritize sentences for fact-checking. 
\\

Figure~\ref{fig:pipeline} shows how the different tasks relate to each other. The first step is to detect tweets that contain check-worthy claims (Task~1; also, Task~5, which is on debates and speeches). The next step is to check whether a target check-worthy claim has been previously fact-checked (Task~2). If not, then there is a need for fact-checking, which involves supporting evidence retrieval (Task~3), followed by actual fact-checking based on that evidence (Task~4).
Tasks~1, 3, and 4 were run for Arabic, while Tasks~1, 2 and 5 were offered for English.

The rest of the paper is organized as follows. Section~\ref{sec:related} 
discusses related work. Section~\ref{sec:ar} describes the tasks that were run in Arabic (Tasks~1, 3 and 4). Section~\ref{sec:en} presents the tasks that were run in English (Tasks~1, 2, and 5). 
Note that Sections~\ref{sec:ar} and~\ref{sec:en} are not exhaustive; the reader should refer to~\cite{clef-checkthat-ar:2020} and~\cite{clef-checkthat-en:2020}, respectively, for further details.
Finally, Section~\ref{sec:conclusions} concludes with final remarks. 

\section{Related Work}
\label{sec:related}


Both the information retrieval and the natural language processing communities have invested significant efforts in the development of systems to deal with disinformation, misinformation, factuality, and credibility.
There has been work on checking the factuality/credibility of a claim, of a news article, or of an information source~\cite{ba2016vera,ACL2020:What:was:written,source:multitask:NAACL:2019,R17-1046,ma2016detecting,mukherjee2015leveraging,popat2016credibility,zubiaga2016analysing}. Claims can come from different sources, but special attention has been paid to those originating in social media~\cite{gupta2014tweetcred,mitra2015credbank,shu2017fake,zhao2015enquiring}. 
Check-worthiness estimation is still a relatively under-explored problem, and has been previously addressed primarily in political debates and speeches~\cite{RANLP2017:debates,Hassan:15,Hassan2016ComparingAF,hassan2017claimbuster,RANLP2019:checkworthiness:multitask}, and only recently in social media~\cite{alam2020fighting}. Similarly, severely under-explored is the task of detecting previously fact-checked claims~\cite{shaar-etal-2020-known}.

\begin{figure}[t]
\centering
\includegraphics[width=\columnwidth]{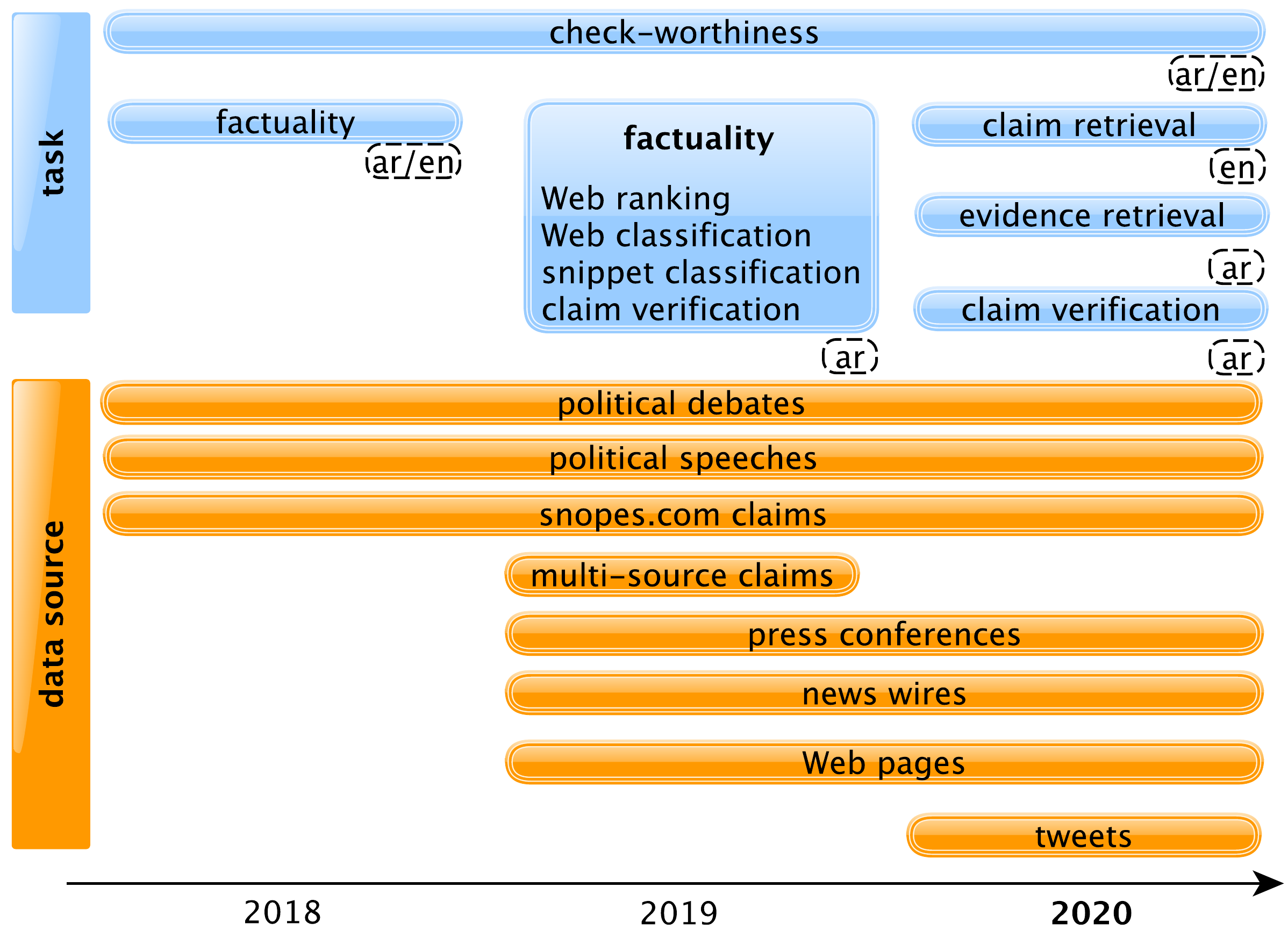}
\caption{The evolution of the tasks at the \ct lab over its three editions. Top:~the unfolding of the tasks that compose a full fact-checking pipeline. Bottom:~the source texts and the genres included in the datasets used by these tasks.}
\label{fig:evolution}
\end{figure}

This is the third edition of the \ct lab, and it represents a clear evolution 
from the tasks that were featured in the previous two editions. Figure~\ref{fig:evolution} 
shows the evolution of the \ct tasks over these three years. The lab started in 
2018 with only two tasks: check-worthiness estimation and factuality (fact-checking), with focus on political debates, speeches, and claims. In that first edition, the English language was leading and the Arabic datasets were produced by translation (manual or automatic with post-editing). The \ct 2019 lab offered a continuity in the check-worthiness task. The Arabic task ---still under the 
factuality umbrella--- started to unfold into four subtasks in order to boost 
the development of models specialized in each of the stages of verification, 
from the ranking of relevant Websites to the final claim verification. 
Regarding data, transcriptions of press conferences were added, as well as news to be used to verify the claims. In 2020, we unfolded the claim verification pipeline into four differentiated tasks. Regarding data, all tasks in 2020 turned to micro-blogging with focus on Twitter, with the exception of legacy task 5 on  check-worthiness, which focused on political debates and speeches. 

Below, we present a brief overview of the tasks and of the most successful 
approaches in the 2019 and the 2018 editions of the lab. 
We refer the reader to~\cite{clef-checkthat:2019} and to~\cite{clef2018checkthat} for more detailed overviews of these earlier editions.

\subsection{\ct 2019}
\label{sub:ct-19}

The 2019 edition of the \ct lab featured two tasks~\cite{clef-checkthat:2019}:

\paragraph{Task~1$_{2019}$.} \emph{Given a political debate, an interview, or 
a speech, transcribed and segmented into sentences, rank the sentences by the 
priority with which they should be fact-checked.}
\\

The most successful approaches used by the participating teams relied on neural networks for the classification of the instances instances. For example, Hansen et 
al.~\cite{T1-Hansen:2019} learned domain-specific word embeddings and syntactic 
dependencies and applied an LSTM classifier. They pre-trained the network with 
previous Trump and Clinton debates, supervised weakly with the ClaimBuster 
system. Some efforts were carried out in order to consider context. Favano et 
al.~\cite{T1-T2-Favano:2019} trained a feed-forward neural network, including 
the two previous sentences as a context. While many approaches relied on 
embedding representations, feature engineering was also popular~\cite{T1-Gasior:2019}. We refer the interested reader to~\cite{clef-checkthat-T1:2019} for further details.

\paragraph{Task~2$_{2019}$.} \emph{Given a claim and a set of 
potentially-relevant Web pages, identify which of the pages (and passages 
thereof) are useful for assisting a human in fact-checking the claim. Finally, 
determine the factuality of the claim.}
\\

The systems for evidence passage identification followed two approaches. BERT 
was trained and used to predict whether an input passage is useful to fact-check 
a claim~\cite{T1-T2-Favano:2019}. Other participating systems used classifiers 
(e.g., SVM) with a variety of features including the similarity between the 
claim and a passage, bag of words, and named entities~\cite{T2Haouari:2019}. As for predicting claim veracity, the most effective approach used a textual 
entailment model. The input was represented using word embeddings and external 
data was also used in training~\cite{T2-Ghanem:2019}. See~\cite{clef-checkthat-T2:2019} for further details.

\medskip
Note that Task~5 of the 2020 edition of the \ct lab is a follow-up of Task~1$_{2019}$, while Task~1 of the 2020 edition is a reformulation that focuses on tweets. In contrast,~Task~2$_{2019}$ was decomposed into two tasks in 2020: Tasks~3 and~4.

\subsection{\ct 2018}
\label{sub:ct-18}

The 2018 edition featured two tasks:

\paragraph{Task~1$_{2018}$} was identical to Task~1$_{2019}$.
\\

The most successful approaches used either a multilayer perceptron or an SVM\@. 
Zuo et al.~\cite{T1-Zuo:2018} enriched the dataset by producing 
\textit{pseudo-speeches} as a concatenation of all interventions by a debater. 
They used averaged word embeddings and bag of words as representations. Hansen 
et al.~\cite{T1-Hansen:2018} represented the entries with embeddings, part of 
speech tags, and syntactic dependencies, and used a GRU neural network with 
attention as a learning model. More details can be found in the task overview paper~\cite{clef-checkthat-T1:2018}.

\paragraph{Task~2$_{2018}$.} \emph{Given a check-worthy claim in the form of a 
(transcribed) sentence, determine whether the claim is likely to be true, 
half-true, or false.}
\\

The best way to address this task was to retrieve relevant information from the 
Web, followed by a comparison against the claim in order to assess its 
factuality. After retrieving such \textit{evidence}, it is fed into the 
supervised model, together with the claim in order to assess its veracity. In 
the case of Hansen et al.~\cite{T1-Hansen:2018}, they fed the claim and the most similar 
Web-retrieved text to convolutional neural networks and SVMs. Meanwhile, Ghanem 
et al.~\cite{T1-Ghanem:2018} computed features, such as the similarity between 
the claim and the Web text, and the Alexa rank for the website.
Once again, this year a similar procedure had to be carried out, but this time 
explicitly decomposed into tasks 3 and 4.
We refer the interested reader to~\cite{clef-checkthat-T2:2018} for further details.

\section{Overview of the Arabic Tasks}
\label{sec:ar}

In order to enable research on Arabic claim verification, we ran three tasks from the verification pipeline (see Figure~\ref{fig:pipeline}) over Arabic tweets. These tasks are check-worthiness on tweets (Task 1), evidence retrieval (Task 3), and claim verification (Task 4). They attracted nine teams. Below, we describe the evaluation dataset created to support each of these tasks. We also present a summary of the approaches used by the participating systems, and we discuss the evaluation results. Further details can be found in our extended overview paper~\cite{clef-checkthat-ar:2020}.

\subsection{Task~1$_{ar}$. Check-Worthiness on Tweets}
\label{sec:ar:t1}

Since check-worthiness estimation for tweets in general, and for \emph{Arabic} tweets in particular, is a relatively new task, we constructed a new dataset specifically designed for training and evaluating systems for this task. We identified the need for a ``context'' that affects check-worthiness of tweets and we used ``topics'' to represent that context. Given a topic, we define a check-worthy tweet as a tweet that is relevant to the topic, contains one main claim that can be fact-checked by consulting reliable sources, and is important enough to be worthy of verification. More on the annotation criteria is presented later in this section. 

\paragraph{\textbf{Dataset}}
To construct the dataset for this task, we first manually created 15 topics over the period of several months. The topics were selected based on trending topics at the time among Arab social media users. Each topic was represented using a title and a description. Some example topic titles include: ``Coronavirus in the Arab World", ``Sudan and normalization", and ``Deal of the century". 
Additionally, we augmented the topic by a set of keywords, hashtags and usernames to track in Twitter. Once we created a topic, we immediately crawled a 1-week stream using the constructed search terms, where we searched Twitter (via Twitter search API) using each term by the end of each day. We limited the search to original Arabic tweets (i.e., we excluded retweets). We de-duplicated the tweets and we dropped tweets matching our qualification filter that excludes tweets containing terms from a blacklist of explicit terms and tweets that contain more than four hashtags or more than two URLs. Afterwards, we ranked the tweets by popularity (defined by the sum of their retweets and likes) and selected the top 500 to be annotated. 

\newcolumntype{g}{>{\columncolor{Gray}}c}
\begin{table}[t]
\caption{Summary of the best approaches for Task~1 Arabic for the participating teams. Shown is information about the learning models (including Transformers), about the main representations, whether the participants used external data, and whether they used machine translation to be able to use additional data from English (MT).}
\label{tab:approaches-ar-1}
\centering
\begin{tabular}{ll | ggggg |ggggg | gggggg | gg}
\toprule
\rowcolor{white}
\multicolumn{2}{l|}{\bf Team} & \multicolumn{5}{c|}{\bf Models}	& \multicolumn{5}{c|}{\bf Distrib.} & \multicolumn{6}{c|}{\bf Represent.}	& \multicolumn{2}{c}{\bf Other}\\
\rowcolor{white}
&&\rotatebox{90}{\bf BERT } &
\rotatebox{90}{\bf Bi--LSTM}	& 
\rotatebox{90}{\bf NN}	& 
\rotatebox{90}{\bf SVM}	& 
\rotatebox{90}{\bf SGD}	& 

\rotatebox{90}{\bf Laser}	& 
\rotatebox{90}{\bf FastText}	& 
\rotatebox{90}{\bf GloVe}	& 
\rotatebox{90}{\bf word2vec}	& 
\rotatebox{90}{\bf PCA}	& 

\rotatebox{90}{\bf One-hot}	& 
\rotatebox{90}{\bf Morphology}	& 
\rotatebox{90}{\bf Syntax}	& 
\rotatebox{90}{\bf Sentiment} & 
\rotatebox{90}{\bf Dependencies}	& 
\rotatebox{90}{\bf NER}	& 
\rotatebox{90}{\bf External data}	& 
\rotatebox{90}{\bf MT}\\
\midrule		 
\rowcolor{white}
Accenture		& \cite{clef-checkthat-williams:2020}		& \faCircle	& 				&			& 			&			& 			&	 			&			& 				& 			&				& 				& 			& 				& 				&			& \faCircle	& \faCircle	\\
bigIR			& \cite{clef-checkthat-Hasanain:2020}		& \faCircle	&				&			& 			&			& 			&	 			&			& 				& 			&				&			 	& 			& 				& 				&			& \faCircle	& 	\\
\rowcolor{white}
Check\_square 	& \cite{clef-checkthat-cheema:2020}			& 			&				&			& \faCircle	&			& 			&	 			&			& \faCircle		& \faCircle	&				&  \faCircle	& 			& 				& \faCircle		&			& 			& 	\\
DamascusTeam		& \cite{clef-checkthat-Hussein:2020}	& 			& \faCircle		&			& 			&			& 			&	 			&			& 				& 			&\faCircle		& 				& 			& 				& 				&			& \faCircle	& 	\\
\rowcolor{white}
EvolutionTeam 	& \cite{clef-checkthat-Touahri:2020}		& 			&				&			& 			&			& 			&	 			&			& 				& 			&				& 				&			& \faCircle		& 				&\faCircle	& \faCircle	& 	\\
NLP\&IR@UNED	& \cite{clef-checkthat-Martinez-Rico:2020}	& 			& 				& \faCircle	& 			&			& 			&	 			& \faCircle	& 				& 			&				& 				& 			& 				& 				&			&			& 	\\
\rowcolor{white}
TOBB ETU		& \cite{clef-checkthat-Kartal:2020}			& \faCircle	&				&			&			&			& 			& \faCircle		&			& 				& 			&				& 				& 			& 				& 				&			&			&	\\
WSSC\_UPF 		& --										& 			&				&			& 			& \faCircle	& \faCircle	&	 			&			& 				& 			&				& \faCircle		& \faCircle	& 				& 				&			& \faCircle	& 	\\

%
\bottomrule
 \end{tabular}
\end{table}

The annotation process was performed in two steps; we first identified the tweets that are relevant to the topic and contain factual claims, then identified the check-worthy tweets among those relevant tweets. 

We first recruited one annotator to annotate each tweet for its relevance to the target topic. In this step, we labeled each tweet as one of three categories:
\begin{itemize}
    \item Non-relevant tweet for the target topic.
    \item Relevant tweet but \emph{with no factual claims}, such as tweets expressing opinions about the topic, references, or speculations about the future, etc.
    \item Relevant tweet that contains a factual claim that can be fact-checked by consulting reliable sources.
\end{itemize}

Only relevant tweets with factual claims were then labelled for check-worthiness. Two annotators initially annotated the relevant tweets. A third \emph{expert} annotator performed disagreement resolution whenever needed. Due to the subjective nature of check-worthiness, we chose to represent the check-worthiness criteria by several questions, to help annotators think about different aspects of check-worthiness. Annotators were asked to answer the following three questions for each tweet (using a scale of 1-5):
\begin{itemize}
    \item Do you think the claim in the tweet is of interest to the public?
    \item To what extent do you think the claim can negatively affect the reputation of an entity, country, etc.?
    \item Do you think journalists will be interested in covering the spread of the claim or the information discussed by the claim? 
\end{itemize}

Once an annotator answers the above questions, she/he is required to answer the following fourth question considering all the ratings given previously: ``Do you think the claim in the tweet is check-worthy?". This question is a yes/no question, and the resulting answer is the label we use to represent check-worthiness in this dataset.

For the final set, all tweets but those labelled as check-worthy were considered not check-worthy. Given 500 tweets annotated for each of the fifteen topics, the annotated set contained 2,062 check-worthy claims (27.5\%). Three topics constituted the training set and the remaining twelve topics were used to later evaluate the participating systems. 

\paragraph{\textbf{Overview of the approaches}} 
Eight teams participated in this task submitting a total of 28 runs. 
Table~\ref{tab:approaches-ar-1} shows an overview of the approaches. The most successful runs adopted fine-tuning existing pre-trained models, namely AraBERT and multilingual BERT models. Other approaches relied on pre-trained models such as Glove, Word2vec, and Language-Agnostic SEntence Representations (LASER) to obtain embeddings for the tweets, which were fed either to neural network models or to traditional machine learning models such as SVMs. In addition to text representations, some teams included other features to their models, namely morphological and syntactic features, part-of-speech (POS) tags, named entities, and sentiment features. 

\paragraph{\textbf{Evaluation}}
We treated Task 1 as a ranking problem where we expected check-worthy tweets to be ranked at the top. We evaluated the runs using precision at $k$ ($P@k$) and Mean Average Precision (MAP). We considered $P@30$ as the official measure, as we anticipated the user would check maximum of 30 claims per week. We also developed two simple baselines: \emph{baseline 1}, which ranks tweets in descending order based on their popularity score (sum of likes and retweets a tweet has received) and \emph{baseline 2}, which ranks tweets in reverse chronological order, i.e., most-recent first. Table~\ref{tab:eval_ar_t1} shows the performance of the best run per team in addition to the two baselines, ranked by the official measure. We can see that most teams managed to improve over the two baselines by a large margin.

\begin{table}[t]
\caption{Performance of the best run per team for Arabic Task~1.}
\label{tab:eval_ar_t1}
\centering
\begin{tabular}{lcccccc}
\toprule
\bf RunID	& \bf $P@10$ & \bf $P@20$	& \bf $P@30$	& \bf MAP	\\
\midrule
Accenture-AraBERT &     0.7167 &     0.6875 &     0.7000 &     0.6232 \\
TOBB-ETU-AF &     0.7000 &     0.6625 &     0.6444 &     0.5816 \\
bigIR-bert &     0.6417 &     0.6333 &     0.6417 &     0.5511 \\
Check\_square-w2vposRun2 &     0.6083 &     0.6000 &     0.5778 &     0.4949 \\
DamascusTeam-Run03 &     0.5833 &     0.5750 &     0.5472 &     0.4539 \\
NLP\&IR@UNED-run4 &     0.6083 &     0.5625 &     0.5333 &     0.4614 \\
baseline2 &     0.3500 &     0.3625 &     0.3472 &     0.3149 \\
baseline1 &     0.3250 &     0.3333 &     0.3417 &     0.3244 \\
EvolutionTeam-Run1 &     0.2500 &     0.2667 &     0.2833 &     0.2675 \\
WSSC\_UPF-RF01 &     0.1917 &     0.1667 &     0.2028 &     0.2542 \\
\bottomrule
\end{tabular}
\end{table}

\subsection{Task~3$_{ar}$. Evidence Retrieval}
\paragraph{\textbf{Dataset}}
For this task, we needed a set of claims and a set of potentially-relevant Web pages from which evidence snippets will be extracted by a system. 

We first collected the set of Web pages using the topics we developed for Task 1. While developing the topics, we represented each one by a set of search phrases. We used these phrases in Google Web search daily as we crawled tweets for the topic. By the end of a week, we collected a set of Web pages that was ready to be used for constructing a dataset to evaluate evidence retrieval systems. 

As for the set of claims, we draw a random sample from the check-worthy tweets identified for each topic for Task 1. Since data from Task 2, Subtask C in the last year's edition of the lab could be used for training~\cite{clef-checkthat-T2:2019}, we only released test claims and Web pages from the twelve test topics used in Task 1. The dataset for this task contains a total of 200 claims and 14,742 corresponding Web pages. 

Since we seek a controlled method to allow systems to return snippets, which in turn would allow us to label a consistent set of potential evidence snippets, we automatically pre-split these pages into snippets that we eventually released per page. 
To extract snippets from the Web pages, we first de-duplicated the crawled Web pages using the page URL\@. Then, we extracted the textual content from the HTML document for each page after removing any markup and scripts. Finally, we detected Arabic text and split it into snippets, where full-stops, question marks, or exclamation marks delimit the snippets. Overall, we extracted 169,902 snippets from the Web pages.\

Due to the large number of snippets collected for the claims, annotating all pairs of claims and snippets was not feasible given the limited time. Therefore, we followed a \emph{pooling} method; we annotate pooled evidence snippets returned from submitted runs by the participating systems. Since the official evaluation measure for the task was set to be $P@10$, we first extracted the top 10 evidence snippets returned by each run for each claim. We then created a pool of unique snippets per claim (considering both snippet IDs and content for de-duplication). Finally, a single annotator annotated each snippet for a claim. The annotators were asked to decide whether a snippet contains evidence useful to verify the given claim. An evidence can be statistics, quotes, facts extracted from verified sources, etc. 

Overall, we annotated 3,380 snippets. After label propagation, we had 3,720 annotated snippets of which only 95 are evidence snippets. Our annotation volume was limited due to the very small number of runs participating in the task (2 runs submitted by one team).

\paragraph{\textbf{Overview of the Approaches}}
One team, EvolutionTeam, submitted two runs for this task~\cite{clef-checkthat-Touahri:2020}. They used machine learning models with two different types of features in each of the runs. In one run, they exploited the similarity feature by computing the cosine similarity between the claim and the snippets to rank them accordingly. They also explored the effectiveness of using linguistic features to rank snippets for usefulness in the second run for which they reported use of external data. 

\paragraph{\textbf{Evaluation}}
This task is modeled as a ranking problem where the system is expected to return evidence at the top of the list of returned snippets. In order to evaluate the submitted runs, we computed $P@k$ at different cutoff ($k$ = 1, 5, 10). The official measure was $P@10$. The team's best-performing run achieved an average $P@10$ of 0.0456 over the claims.

\subsection{Task~4$_{ar}$. Claim Verification}
Starting with the same 200 claims used in Task 3, one expert fact-checker verified each claim's veracity. We limited the annotation categories to two, true and false, excluding partially-true claims. A true claim is a claim that is supported by a reliable source that confirms the authenticity of the information published in the tweet. A false claim can be a claim that mentions information contradicting that in a reliable source or has been explicitly refuted by a reliable source.

\paragraph{\textbf{Dataset}}
The claims in the tweets were annotated considering two main factors; the content of the tweet (claim) and the date of the tweet publication. For the annotation, we considered supporting or refuting information that was reported before, on, or few days after the time of the claim. We consulted several reliable sources to verify the claims. The sources that were used differed according to the topic of the claim. For example, for health-related claims, we consulted refereed studies or articles published in reliable medical journals or websites such as APA\footnote{\url{https://www.apa.org/}}.

Out of the initial 200 claims, we ended up with 165 claims for which we managed to find a definite label. Six claims among these 165 were found to be False. Since data from Task 2, Subtask D in the last year's edition of the lab can be used for training~\cite{clef-checkthat-T2:2019}, the final set of 165 annotated claims was used to evaluate the submitted runs.

\paragraph{\textbf{Evaluation}}
For this task, there were a total of two runs submitted by the same team, EvolutionTeam. The models relied on linguistic features, and they used external data in one of the runs. We treated the task as a classification problem and we used typical evaluation measures for such tasks in the case of class imbalance: Precision, Recall, and F1 score. The latter was the official evaluation measure. The best-performing run achieved a macro-averaged F1 score of 0.5524.

\section{Overview of the English Tasks}
\label{sec:en}
This year we proposed three of the tasks of the verification pipeline in 
English: check-worthiness estimation over tweets, verified claim retrieval, 
and check-worthiness estimation in political debates and speeches
(cf.\ Figure~\ref{fig:pipeline}).
A total of 18 teams participated in the English tasks.

\subsection{Task~1$_{en}$. Check-Worthiness on Tweets}

\paragraph{\textbf{Task 1 (English)}}\textit{Given a topic and a stream of potentially-related tweets, rank the tweets according to their check-worthiness for the topic.} 
\\

Previous work on check-worthiness focused primarily on political debates and speeches, while here we focus on tweets instead.

\paragraph{\textbf{Dataset}}
We focused on a single topic, namely \emph{COVID-19}, and we collected tweets that matched one of the following keywords and hashtags: \emph{\#covid19, \#CoronavirusOutbreak, \#Coronavirus, \#Corona, \#CoronaAlert, \#CoronaOutbreak, Corona}, and \emph{covid-19}.
We ran all the data collection in March 2020, and we selected the most retweeted tweets for manual annotation.
\medskip

For the annotation, we considered a number of factors. These include tweet popularity in terms of retweets, which is already taken into account as part of the data collection process. We further asked the annotators to answer the following five questions:\footnote{We used the following MicroMappers setup for the annotations:\\ \url{http://micromappers.qcri.org/project/covid19-tweet-labelling/}}

\begin{itemize}
    \item \textbf{Q1: Does the tweet contain a verifiable factual claim?} This is an objective question. Positive examples include\footnote{This is influenced by \cite{DBLP:journals/corr/abs-1809-08193}.} tweets that state a definition, mention a quantity in the present or the past, make a verifiable prediction about the future, reference laws, procedures, and rules of operation, discuss images or videos, and state correlation or causation, among others. 
    \item \textbf{Q2: To what extent does the tweet appear to contain false information?} This question asks for a subjective judgment; it does not ask for annotating the actual factuality of the claim in the tweet, but rather whether the claim appears to be false.
    \item \textbf{Q3: Will the tweet have an effect on or be of interest to the general public?}
    This question asks for an objective judgment. Generally, claims that contain information related to potential cures, updates on number of cases, on measures taken by governments, or discussing rumors and spreading conspiracy theories should be of general public interest.
    \item \textbf{Q4: To what extent is the tweet harmful to the society,  person(s), company(s) or product(s)?} This question also asks for an objective judgment: to identify tweets that can negatively affect society as a whole, but also specific person(s), company(s), product(s).
    \item \textbf{Q5: Do you think that a professional fact-checker should verify the claim in the tweet?} This question asks for a subjective judgment. Yet, its answer should be informed by the answer to questions Q2, Q3 and Q4, as a check-worthy factual claim is probably one that is likely to be false, is of public interest, and/or appears to be harmful. 
\end{itemize}

For the purpose of the task, we consider as worth fact-checking the tweets that received a positive answer both to Q1 and to Q5; if there was a negative answer for either Q1 or Q5, the tweet was considered not worth fact-checking. The answers to Q2, Q3, and Q4 were not considered directly, but they helped the annotators make a better decision for Q5.

The annotations were performed by 2--5 annotators independently, and then consolidated after a discussion for the cases of disagreement. The annotation setup was part of a broader COVID-19 annotation initiative; see~\cite{alam2020fighting} for more details about the annotation instructions and setup. 

Table~\ref{table:task1-en-stats} shows statistics about the data, which is split into training, development, and testing. We can see that the data is fairly balanced with the check-worthy claims making 34-43\% of the examples across the datasets.

\begin{table*}[t]
  \small
  \centering
  \caption{\label{table:task1-en-stats}\textbf{Task 1, English:} Statistics about the tweets in the dataset.}
  \begin{tabular}{p{0.2\linewidth}cc}
    \toprule
    \bf Partition & \bf Total & \bf Check-worthy \\
    \midrule
    Train & 672 & 231 \\
    Dev & 150 & 59 \\
    Test & 140 & 60 \\
    \bottomrule
  \end{tabular}
\end{table*}


\begin{table}[h]
\caption{\label{tab:approaches-en-1}\textbf{Task 1, English:} Summary of the approaches used in the primary system submissions. Shown is which systems used transformers, learning models, distributional features, standard features, and other.
}
\centering
\begin{tabular}{lc | ggg | gggggg |gggg | ggggggg | gg}
\toprule

\rowcolor{white}
\multicolumn{2}{l|}{\bf Team} & \multicolumn{3}{c|}{\bf Transf}	& \multicolumn{6}{c|}{\bf Models}	& \multicolumn{4}{c|}{\bf Distrib.} & \multicolumn{4}{c|}{\bf Features}	& \multicolumn{2}{c}{\bf Other}\\
\rowcolor{white}
&&
\rotatebox{90}{\bf BERT } &
\rotatebox{90}{\bf RoBERTa}	& 
\rotatebox{90}{\bf Huggingface}	& 
\rotatebox{90}{\bf BiLSTM}	& 
\rotatebox{90}{\bf CNN}	& 
\rotatebox{90}{\bf Rnd forest}	& 
\rotatebox{90}{\bf Linear reg}	&
\rotatebox{90}{\bf Logistic reg}	& 
\rotatebox{90}{\bf SVM}	& 

\rotatebox{90}{\bf FastText}	& 
\rotatebox{90}{\bf GloVe}	& 
\rotatebox{90}{\bf PCA}	& 
\rotatebox{90}{\bf Topic models}	& 

\rotatebox{90}{\bf tf--idf}	& 
\rotatebox{90}{\bf Dependencies} & 
\rotatebox{90}{\bf POS}	& 
\rotatebox{90}{\bf NEs}	& 
\rotatebox{90}{\bf Ext.\ data}	& 
\rotatebox{90}{\bf Graph relations}
\\
\midrule		 
\rowcolor{white}
Accenture		& \cite{clef-checkthat-williams:2020}		& 			& \faCircle	& 			& 			& 			& 			& 			& 			& 			& 			& 			& 			& 			& 			& 			& 			& 			& 			& 	  \\
BustingMisinformation	& --								& 			& 			& 			& 			& \faCircle	& 			& 			& 			& \faCircle	& 			& \faCircle	& 			& \faCircle	& \faCircle	& 			& 			& 			& 		& 	  \\
\rowcolor{white}
Check\_square	& \cite{clef-checkthat-cheema:2020}		& \faCircle	& 			& 			& 			& 			& 			& 			& 			& \faCircle	& 			& 			& \faCircle	& 			& 			& \faCircle	& \faCircle	& \faCircle	& 		& 	  \\
Factify			& --										& \faCircle	& 			& 			& 			& 			& 			& 			& 			& 			& 			& 			& 			& 			& 			& 			& 			& 			& \faCircle	& 	  \\
\rowcolor{white}
NLP\&IR@UNED	& \cite{clef-checkthat-Martinez-Rico:2020}	& 			& 			& 			& \faCircle	& 			& 			& 			& 			& 			& 			& \faCircle	& 			& 			& 			& 			& 			& 			& 		& \faCircle	\\
QMUL-SDS		& \cite{clef-checkthat-Alkhalifa:2020}		& \faCircle	& 			& 			& 			& \faCircle	& 			& 			& 			& 			& 			& 			& 			& 			& 			& 			& 			& 			& 		& 	  \\
\rowcolor{white}
Team\_Alex		& \cite{clef-checkthat-Nikolov:2020}		& 			& \faCircle			& 			& 			& 			& 			& 			& 			& 			& 			& 			& 			& 			& 			& 			& 			& 			& 		& 	 \\
TheUofSheffield	& \cite{clef-checkthat-McDonald:2020}		& 			& 			& 			& 			& 			& \faCircle	& 			& 			& 			& \faCircle	& 			& 			& 			& \faCircle	& 			& 			& 			& 		& 	 \\
\rowcolor{white}
TOBB ETU		& \cite{clef-checkthat-Kartal:2020}			& \faCircle	& 			& 			& 			& 			& 			& 			& \faCircle	& 			& 			& 			& 			& 			& 			& 			& \faCircle	& 			& 		& 	  \\
UAICS 			& --										& \faCircle	& 			& \faCircle	& 			& 			& 			& 			& 			& 			& 			& 			& 			& 			& 			& 			& 			& 			& 		& 	 \\		
\rowcolor{white}
SSN\_NLP		& --										& 			& \faCircle	& 			& 			& 			& 			& 			& 			& 			& 			& 			& 			& 			& 			& 			& 			& 			& 		& 	  \\
ZHAW			& --										& 			& 			&			& 			& 			& 			& \faCircle	& 			& 			& 			& 			& 			& 			& 			& 			& \faCircle	& \faCircle	& 		& 	\\		

\bottomrule
\end{tabular}
\end{table}

\begin{table}[h]
\caption{\label{tab:eval_en_t1}\textbf{Task 1, English:} Evaluation results for the primary submissions.}
\centering
\begin{tabular}{lccccccccc}
\toprule 
\bf Team			& \bf MAP	& \bf RR	& \bf R-P	& \bf P@1	& \bf P@3	& \bf P@5	& \bf P@10	& \bf P@20	& \bf P@30    \\
\midrule
Accenture			& \bf 0.806	& \bf 1.000	& \bf 0.717	& \bf 1.000	& \bf 1.000	& \bf 1.000	& \bf 1.000	& \bf 0.950	& \bf 0.740  \\
Team\_Alex			& 0.803		& \bf 1.000	& 0.650		& \bf 1.000	& \bf 1.000	& \bf 1.000	& \bf 1.000	& \bf 0.950	& \bf 0.740   \\
Check\_square		& 0.722		& \bf 1.000	& 0.667		& \bf 1.000	& 0.667		& 0.800		& 0.800		& 0.800	 	& 0.700   \\
QMUL-SDS			& 0.714		& \bf 1.000	& 0.633		& \bf 1.000	& \bf 1.000	& \bf 1.000	& 0.900	& 0.800	 & 0.640   \\
TOBB ETU			& 0.706		& \bf 1.000	& 0.600		& \bf 1.000	& \bf 1.000	& \bf 1.000	& 0.900	& 0.800	 & 0.660   \\
SSN\_NLP			& 0.674		& \bf 1.000	& 0.600		& \bf 1.000	& \bf 1.000	& 0.800	& 0.800	& 0.800	& 0.620   \\
Factify				& 0.656		& 0.500		& 0.683		& 0.000	& 0.333	& 0.600	& 0.700	& 0.750	& 0.700   \\
BustingMisinformation& 0.617	& \bf 1.000	& 0.583		& \bf 1.000	& \bf 1.000	& 0.800	& 0.700	 & 0.600	& 0.600   \\
NLP\&IR@UNED		& 0.607		& \bf 1.000	& 0.567		& \bf 1.000	& \bf 1.000	& \bf 1.000	& 0.700	& 0.600	& 0.580   \\
\it Baseline ($n$-gram)	& 0.579		& 1.000		& 0.500		& 1.000	&	0.667	&	0.800	&	0.800	&	0.700	&	0.600   \\
ZHAW				& 0.505		& 0.333		& 0.533		& 0.000		& 0.333	& 0.400	& 0.600	& 0.500	& 0.520   \\
UAICS				& 0.495		& \bf 1.000	& 0.467		& \bf 1.000	& 0.333	& 0.400	& 0.600	& 0.600	& 0.460   \\
TheUofSheffield		& 0.475		& 0.250		& 0.533		& 0.000		& 0.000	& 0.400	& 0.200	& 0.350	& 0.480   \\
\bottomrule
\end{tabular}
\end{table}

\paragraph{\textbf{Evaluation}}
This is a ranking task, where a tweet has to be ranked according to its 
check-worthiness. Therefore, we consider mean average precision (MAP) as the official evaluation measure, which we complement with reciprocal rank (RR), R-precision (R-P), and P@$k$ for $k \in \{1, 3, 5, 10, 20, 30\}$.
The data and the evaluation scripts are available online.\footnote{\url{https://github.com/sshaar/clef2020-factchecking-task1/}}

\paragraph{\textbf{Overview of the approaches}}
A total of 12 teams took part in Task 1. The submitted models range from state-of-the-art Transformers such as BERT and RoBERTa to more traditional machine learning models such as SVM and Logistic Regression. Table~\ref{tab:approaches-en-1} shows a summary of the approaches used by the primary submissions of the participating teams. The highest overall score was achieved using a RoBERTa model.

The top-ranked team \textbf{Accenture} used RoBERTa with mean pooling and dropout.

The second-best \textbf{Team\_Alex} trained a logistic regression classifier using as features the RoBERTa\'s cross-validation predictions on the data and metadata from the provided JSON file as features.

Team \textbf{Check\_square} used BERT embeddings along with syntactic features with SVM\slash PCA and ensembles.

Team \textbf{QMUL-SDS} fine-tuned the uncased COVID-Twitter-BERT architecture, which was pre-trained on COVID-19 Twitter stream data. 

Team \textbf{TOBB ETU} used BERT and word embeddings as features in a logistic regression model, adding POS tags and important hand-crafted word features.

Team \textbf{SSN\_NLP} also used a RoBERTa classifier.

Team \textbf{Factify} submitted a BERT-based classifier.

Team \textbf{BustingMisinformation} used an SVM with TF-IDF features and GloVe embeddings, along with topic modelling using NMF.

Team \textbf{NLP\&IR@UNED} trained a bidirectional LSTM on top of GloVe embeddings. They increased the number of inputs with a graph generated from the additional information provided for each tweet.

Team \textbf{ZHAW} used a logistic regression with POS tags and named entities along with additional features about the location of posting, its time, etc.

Team \textbf{UAICS} submitted predictions from a fine-tuned custom BERT large model.

Team \textbf{TheUofSheffield} trained a custom 4-gram FastText model. Their pre-processing includes lowercasing, lemmatization, as well as URL, emoji, stop words, and punctuation removal.


Table~\ref{tab:eval_en_t1} shows the performance of the primary submissions to Task~1 in English. We can see that Accenture and Team\_Alex achieved very high scores on all evaluation measures and outperformed the remaining teams by a wide margin, e.g.,~by about eight points absolute in terms of MAP\@. We can further see that most systems managed to outperform an $n$-gram baseline by a very sizeable margin.

\subsection{Task~2$_{en}$. Verified Claim Retrieval}

\paragraph{\textbf{Task 2 (English)}}\textit{Given a check-worthy input claim and a set of verified claims, rank those verified claims, so that the claims that can help verify the input claim, or a sub-claim in it, are ranked above any claim that is not helpful to verify the input claim..} 
\\

Unlike the other tasks of the \ctend lab, Task~2 is a new one.
Table~\ref{table:task2example} shows an example of a tweet by Donald Trump claiming that a video footage about Syria aired by BBC is fake (input claim), and it further shows some already verified claims ranked by their relevance with respect to the input claim. 

Note that the input claim and the most relevant verified claim, while expressing the same concept, are phrased quite differently. 
A good system for ranking the verified claims might greatly reduce the time that a fact-checkers or a journalist would need to check whether a given input claim has already been fact-checked. 

\begin{table}[ht]
    \small
    \centering
    \caption{\label{table:task2example}\textbf{Task 2, English:} example input. A subset of verified claims ordered by relevance with respect to the input claim according to our baseline system.}
    \begin{tabular}{p{1.2cm}cp{10.5cm}}
        \toprule
        \textbf{input tweet}: & & A big scandal at @ABC News. They got caught using really gruesome FAKE footage of the Turks bombing in Syria. A real disgrace. Tomorrow they will ask softball questions to Sleepy Joe Biden’s son, Hunter, like why did Ukraine \& China pay you millions when you knew nothing? Payoff? — Donald J. Trump (@realDonaldTrump) October 15, 2019\\
        \toprule
        \textbf{verified claims}: & \bf (1) & ABC News mistakenly aired a video from a Kentucky gun range during its coverage of Turkey's attack on northern Syria in October 2019. \\
        & \bf (2) & In a speech to U.S. military personnel, President Trump said if soldiers were real patriots, they wouldn't take a pay raise.\\
        & \bf (3) & Former President Barack Obama tweeted: ``Ask Ukraine if they found my birth certificate.'' \\
        \toprule
    \end{tabular}
\end{table}


Each input claim was retrieved from the fact-checking website Snopes,\footnote{\url{www.snopes.com}} which dedicates an article to assessing the truthfulness of each claim they have analyzed. In that article, there might be listed different tweets that contain (a paraphrase of) the target claim. Together with the title of the article page and the rating of the claim, as assigned by Snopes, we collect all those tweets and we use them as input claims. Then, the task is, given such a tweet, to find the corresponding claim. The set of target claims consists of the claims that correspond to the tweets we collected, augmented with all Snopes claims collected by ClaimsKG~\cite{ClaimsKG}. 
Note that we have just one list of verified claims, which is used for matching by all input tweets. 

Our data consists of 1,197 input tweets, which we split into training (800 input tweets), development (197 tweets), and test set (200 tweets). 
These input tweets are to be matched against a set of 10,375 verified claims. 




\paragraph{\bf Overview of the approaches} 

A total of eight teams participated in Task 2. 
A variety of scoring functions have been tested, based on supervised learning such as BERT and its variants and SVM, to unsupervised approaches such as simple cosine similarity and scores produced by Terrier and Elastic Search. 
Two teams focused also on data cleaning by removing URLs, hashtags, usernames and emojis from the tweets. 
Table~\ref{tab:approaches-en-2} shows a summary of the approaches used by the primary submissions of the participating teams.

\begin{table}[tbh]
\caption{\label{tab:approaches-en-2}\textbf{Task 2, English:} summary of the approaches used by the primary system submissions. We report which systems used search engines scores, scoring functions (supervised or not), representations (other than Transformers), and the removal of tokens. We further indicate whether external data was used.}
\centering
\begin{tabular}{lc | gg | ggggggg | ggg | ggggg | g}
\toprule

\rowcolor{white}
\multicolumn{2}{l|}{\bf Team} & \multicolumn{2}{c|}{\bf Engine}	& \multicolumn{7}{c|}{\bf Scoring}	& \multicolumn{3}{c|}{\bf Repr.} &  \multicolumn{5}{c|}{\bf Removal}\\
\rowcolor{white}
&&
\rotatebox{90}{\bf Terrier }			&
\rotatebox{90}{\bf ElasticSearch}		& 

\rotatebox{90}{\bf LambdaMART }			&
\rotatebox{90}{\bf BERT } 				&
\rotatebox{90}{\bf RoBERTa}				& 
\rotatebox{90}{\bf Unspecified Transf.}	& 
\rotatebox{90}{\bf KD search}	&
\rotatebox{90}{\bf SVM}					& 
\rotatebox{90}{\bf Cosine}				& 
 
\rotatebox{90}{\bf tf--idf}				& 
\rotatebox{90}{\bf BM25}				& 
\rotatebox{90}{\bf Term dependencies}	& 

\rotatebox{90}{\bf URL removal}			&
\rotatebox{90}{\bf Emoji removal}		&
\rotatebox{90}{\bf Time removal}		&
\rotatebox{90}{\bf Username removal}	&
\rotatebox{90}{\bf Hashtag removal}		&
\rotatebox{90}{\bf External data}
\\
\midrule		 
\rowcolor{white}
Buster.AI		& \cite{clef-checkthat-Bouziane:2020}	& 		& 		& 		& 		& \faCircle	& 		& 	&  			& 			& 			& 			& 			& 	  		& 			& 			& 			& 		& \faCircle	\\
Check\_square		& \cite{clef-checkthat-cheema:2020}	& 		& 		& 		& 		& 		& \faCircle	& \faCircle	& 			& 			& 			& 			& 			& 	  		& 			& 			& 			& 			&\\
\rowcolor{white}
elec-dlnlp		& --					& 		& 		& 		& \faCircle	& 		& 		& 	&  			& 			& 			& \faCircle	& 			& 	  		& 			& 			& 			& 			&\\
iit 			& --					& 		& 		& 		& \faCircle	& 		& 		& 	&  			& \faCircle	& 			& 			& 			& \faCircle	& \faCircle & \faCircle	& 			& 			&\\
\rowcolor{white}
TheUofSheffield		& \cite{clef-checkthat-McDonald:2020}	& 		& 		& 		& 		& 		& 		& 	& \faCircle	& 			& \faCircle	& \faCircle	& 			& \faCircle	& 			& 			& \faCircle	& \faCircle	&\\
trueman 		& --					& 		& 		& 		& 		& 		& \faCircle	& 	&  			& 			& 			& 			& 			& 	  		& 			& 			& 			& 			&\\
\rowcolor{white}
UB\_ET 			& \cite{clef-checkthat-Thuma:2020}	& \faCircle	& 		& \faCircle	& 		& 		& 		& 	&  			& 			& \faCircle	& \faCircle	& \faCircle	& 	  		& 			& 			& 			& 			&\\
UNIPI-NLE		& \cite{clef-checkthat-Passaro:2020}	& 		& \faCircle	& 		& \faCircle	& 		& 		& 	&  			& \faCircle	& 			& 			& 			& 	  		& 			& 			& 			& 			&\\
\bottomrule
\end{tabular}
\end{table}

\begin{table}[tbh]
\caption{\textbf{Task 2, English:} performance for the primary submissions and for an Elastic Search (ES) baseline. }
\label{tab:eval_en_t2}
\centering
\begin{tabular}{+l|^c^c^c^c | ^c^c^c| ^c^c^c}
\toprule 
			& \multicolumn{4}{c|}{\bf MAP}											& \multicolumn{3}{c|}{\bf Precision}								&	\multicolumn{3}{c}{\bf RR}\\
\bf Team		& \bf @1& \bf @3& \bf @5& \bf --& \bf @1	& \bf @3	& \bf @5	& \bf @1	& \bf @3	& \bf @5 \\
\midrule
\rowstyle{\bfseries}%
Buster.AI		& 0.897	& 0.926	& 0.929	& 0.929	& 0.895	& 0.320	& 0.195	&  0.895	& 0.923	& 0.927	\\

UNIPI-NLE		& 0.877	& 0.907	& 0.912	& 0.913	& 0.875	& 0.315	& 0.193	& 0.875	& 0.904	& 0.909	\\
UB\_ET			& 0.818	& 0.862	& 0.864	& 0.867	& 0.815	& 0.307	& 0.186	& 0.815	& 0.859	& 0.862	\\
NLP\&IR@UNED	& 0.807	& 0.851	& 0.856	& 0.861	& 0.805	& 0.300	& 0.185	& 0.805	& 0.848	& 0.854	\\
TheUofSheffield	& 0.807	& 0.807	& 0.807	& 0.807	& 0.805	& 0.270	& 0.162	& 0.805	& 0.805	& 0.805	\\
trueman			& 0.743	& 0.768	& 0.773	& 0.782	& 0.740	& 0.267	& 0.164	& 0.740	& 0.766	& 0.771	\\
elec-dlnlp		& 0.723	& 0.749	& 0.760	& 0.767	& 0.720	& 0.262	& 0.166	& 0.720	& 0.747	& 0.757	\\
Check\_square	& 0.652	& 0.690	& 0.695	& 0.706	& 0.650	& 0.247	& 0.152	& 0.650	& 0.688	& 0.692	\\
\it baseline (ES)	& 0.470	& 0.601	& 0.609	& 0.619	& 0.472 & 0.249 & 0.156 & 0.472	& 0.603	& 0.611	\\
iit				& 0.263	& 0.293	& 0.298	& 0.311	& 0.260	& 0.112	& 0.071	& 0.260	& 0.291	& 0.295	\\
\bottomrule
\end{tabular}
\end{table}

The winning team, \textbf{Buster.AI}, cleaned the tweets from non-readable input and used a pre-trained and fine-tuned version of RoBERTa to build their system. 

Team \textbf{UNIPI-NLE} performed two cascade fine-tunings of a sentence-BERT model. 
Initially, they fine-tuned on the task of predicting the cosine similarity for tweet--claim. For each tweet, they trained on 20 random negative verified claims and the gold verified claim.
The second fine-tuning step fine-tuned the model as a classification task for which sentence-BERT has to output 1 if the pair is correct, and 0 otherwise. They selected randomly two negative examples and used them with the gold to fine-tune the model. 
Before inference, they pruned the verified claim list, top-2500 using Elastic Search and simple word matching techniques. 

Team \textbf{UB\_ET} trained therir model on a limited number of tweet--claim pairs per tweet. They retrieved the top-1000 tweet--claim pairs for each tweet using the DPH information retrieval weighing model and computed several query-related features and then built a LambdaMart model on top of them. 

Team \textbf{NLP\&IR@UNED} used the Universal Sentence Encoder to obtain embeddings for the tweets and for the verified claims. They then trained a feed-forward neural network using the cosine similarity between a tweet and a verified claim, and statistics about the use of words from different parts of speech.

Team \textbf{TheUniversityofSheffield} pre-processed the input tweets in order to eliminate hashtags, and then trained a Linear SVM using as features TF.IDF-weighted cosine similarity and BM25 matching scores between the tweets and the verified claims.

Teams \textbf{trueman} and \textbf{elec-dlnlp} prepared the input tweets to eliminate hashtags and then used Transformer-based similarity along with Elastic Search scores.

Team \textbf{Check\_square} fine-tuned sentence-BERT with mined triplets and KD-search.

Team \textbf{iit} used cosine similarity using a pre-trained BERT model between the embeddings of the tweet and of the verified claim.

\paragraph{\bf Evaluation}

The official evaluation measure for Task~2 is MAP@$k$ for $k=5$. However, we further report MAP for $k \in \{1, 3, 10, 20 \}$, overall MAP, R-Precision, Average Precision, Reciprocal Rank, and Precision@$k$. Table~\ref{tab:eval_en_t2} shows the evaluation results in terms of some of the performance measures for the primary submissions to Task~2.
We can see that the winner Buster.AI and the second-best UNIPI-NLE are well ahead of the remaining teams by several points absolute on all evaluation measures.
We can further see that most systems managed to outperform an Elastic Search (ES) baseline by a huge margin. The data and the evaluation scripts are available online.\footnote{\url{https://github.com/sshaar/clef2020-factchecking-task2/}}

\subsection{Task~5$_{en}$. Check-Worthiness on Debates}

Task~5 is a legacy task that has evolved from the first edition of the \ct lab. In each edition, more data from more diverse sources have been added, always with focus on politics.
The task focuses on mimicking the selection strategy that fact-checkers, e.g.,~in PolitiFact, use to select the sentences and the claims to fact-check. The task is defined as follows:

\paragraph{\textbf{Task 5 (English)}}\textit{Given a transcript, rank the sentences in the transcript according to the priority to fact-check them.} \\

We used \politifact as the main fact-checking source. On \politifact, often after a major political event such as a public debate or a speech by a government official, a journalist would go through the transcript of the event and would select few claims that would then be fact-checked. These claims would then be discussed in an article about the debate, published on the same site.  We collected all such articles, we further obtained the official transcripts of the event from ABCNews, Washington Post, CSPAN, etc. 
Since sometimes the claims published in the articles are paraphrased, we double-checked and we manually matched them to the transcripts.

We collected a total of 70 transcripts, and we annotated them based on overview articles from \politifact. The transcripts belonged to one of four types of political events: Debates, Speeches, Interviews, and Town-halls. We used 50 transcripts for training and 20 for testing. We used the older transcripts for training and the more recent ones for testing. Table~\ref{table:task5-en-stats} shows the total number of sentences of the transcripts and the number of sentences that were fact-checked by \politifact.

\begin{table*}[tbh]
  \small
  \centering
  \caption{\textbf{Task 5, English:} total number of sentences and number of sentences containing claims that are worth fact-checking --- organized by type.}
  \begin{tabular}{l c c c c}
    \toprule
    \bf Type & \bf Partition & \bf Transcripts & \bf Sentences & \bf Check-worthy\\
    \midrule
    Debates & \shortstack{Train \\ Test} & \shortstack{18 \\ 7} & \shortstack{25,688 \\ 11,218} & \shortstack{254 \\ 56} \\
    \midrule
    Speeches & \shortstack{Train \\ Test} & \shortstack{18 \\ 8} & \shortstack{7,402 \\ 7,759} & \shortstack{163 \\ 50} \\
    \midrule
    Interviews & \shortstack{Train \\ Test} & \shortstack{11 \\ 4} & \shortstack{7,044 \\ 2,220} & \shortstack{62 \\ 23} \\
    \midrule
    Town-halls & \shortstack{Train \\ Test} & \shortstack{3 \\ 1} & \shortstack{2,642 \\ 317} & \shortstack{8 \\ 7} \\
    \midrule
    \bf Total & \shortstack{\bf Train \\ \bf Test} & \shortstack{\bf 50 \\ \bf 20} & \shortstack{\bf 42,776 \\ \bf 21,514} & \shortstack{\bf 487 \\ \bf 136} \\
    \bottomrule
  \end{tabular}
  \label{table:task5-en-stats}
\end{table*}

\paragraph{\textbf{Overview of the approaches}}
Three teams participated in this task submitting a total of eight runs. Each of the teams used different text embedding models for the transcripts.
The best results were obtained using GloVe's embeddings. 

Team \textbf{NLPIR01} used 6B-100D GloVe embeddings as an input to a bidirectional LSTM\@. They further tried sampling techniques but without success.

Team \textbf{UAICS} used the TF.IDF representations using sentences unigrams. They then trained different binary classifiers, such as  Logistic regression, Decision Trees, and Na\"{i}ve Bayes, and they found the latter to perform best.

Team \textbf{TOBB ETU P} tried fine-tuning BERT and modeling the task as a classification task, but ultimately used Part-Of-Speech (POS) tags with logistic regression and a handcrafted word list from the dataset as their official submission. 




\paragraph{\textbf{Evaluation}}
As this task was very similar to Task~1, but on a different genre, we used the same evaluation measures: MAP as the official measure, and we also report P@$k$ for various values of $k$. Table~\ref{tab:eval_en_t5} shows the performance of the primary submissions of the participating teams. The overall results are quite low, and only one team managed to beat the $n$-gram baseline. Once again, the data and the evaluation scripts are available online.\footnote{\url{https://github.com/sshaar/clef2020-factchecking-task5/}}

\begin{table}[t]
\caption{Performance of the primary submissions to Task~5 English. }
\label{tab:eval_en_t5}
\centering
\begin{tabular}{l| c| cc | cccccc }
\toprule
\bf Team	& \bf MAP	& \bf RR	& \bf R-P	& \bf P@1	& \bf P@3	& \bf P@5	& \bf P@10	& \bf P@20	& \bf P@30	\\
\midrule
NLPIR01		& \bf 0.087	& \bf 0.277	& \bf 0.093	& \bf 0.150	& \bf 0.117	& \bf 0.130	& \bf 0.095	& \bf 0.073	& \bf 0.039	\\
\it Baseline ($n$-gram) & 0.053	& 0.151	& 0.053	& 0.050 & 0.033 & 0.040 & 0.055	& 0.043	& 0.038	\\
UAICS		&  0.052	&  0.225	&  0.053	& \bf 0.150	&  0.100	&  0.070	&  0.050	&  0.038	&  0.027	\\
TOBB ETU P	&  0.018	&  0.033	&  0.014	&  0.000	&  0.017	&  0.020	&  0.010	&  0.010	&  0.006	\\
\bottomrule
\end{tabular}
\end{table}

\section{Conclusion and Future Work}
\label{sec:conclusions}

We have described the 2020 edition of the \ct lab, intended to foster the creation 
of technology for the (semi-)automatic identification and verification of claims in social media. The task attracted submissions from 23 teams (up from 14 at CLEF 2019): 18 made submissions for English, and 8 for Arabic.
We believe that the technology developed to address the five 
tasks we have proposed will be useful not only as a supportive technology for investigative journalism, but also for the lay citizen, which today needs 
to be aware of the factuality of the information available online.

\section*{Acknowledgments}
This work was made possible in part by NPRP grant\# NPRP11S-1204-170060 from the Qatar National Research Fund (a member of Qatar Foundation). The statements made herein are solely the responsibility of the authors. 
The work of Reem Suwaileh was supported by GSRA grant\# GSRA5-1-0527-18082 from the Qatar National Research Fund and the work of Fatima Haouari was supported by GSRA grant\# GSRA6-1-0611-19074 from the Qatar National Research Fund.
This research is also part of the Tanbih project, which aims to limit the effect of disinformation, ``fake news'', propaganda, and media bias.

\bibliographystyle{splncs04}
\bibliography{sigprop,clef18_checkthat,clef19_checkthat,clef20_checkthat}

\end{document}